\documentclass[runningheads]{llncs}
\usepackage[T1]{fontenc}
\usepackage{graphicx}
\usepackage{booktabs}
\usepackage[table]{xcolor}  
\usepackage{multirow}
\usepackage{caption}
\usepackage{amsmath,amssymb}
\usepackage{subfigure}  
\usepackage{url}
\usepackage{adjustbox}

\definecolor{lightgray}{RGB}{232, 232, 232} 

\begin{document}
\title{SAMed-2: Selective Memory Enhanced Medical Segment Anything Model}
\titlerunning{Selective Memory Enhanced Medical Segment Anything Model}
%
%

\author{Zhiling Yan\inst{1} \and
Sifan Song\inst{2} \and
Dingjie Song\inst{1} \and
Yiwei Li\inst{3} \and
Rong Zhou\inst{1} \and
Weixiang Sun\inst{4} \and
Zhennong Chen\inst{2} \and
Sekeun Kim\inst{2} \and
Hui Ren\inst{2} \and
Tianming Liu\inst{3} \and
Quanzheng Li\inst{2} \and
Xiang Li\inst{2} \and
Lifang He\inst{1} \and
Lichao Sun\inst{1}
} 

\authorrunning{Z. Yan et al.}

\institute{
Lehigh University, Bethlehem, PA, USA 
\and
Massachusetts General Hospital and Harvard Medical School, Boston, MA, USA 
\and
University of Georgia, Athens, GA, USA 
\and
University of Notre Dame, Notre Dame, IN, USA
}

\maketitle              
%

\begin{abstract}
Recent ``segment anything'' efforts show promise by learning from large-scale data, but adapting such models directly to medical images remains challenging due to the complexity of medical data, noisy annotations, and continual learning requirements across diverse modalities and anatomical structures. In this work, we propose SAMed-2, a new foundation model for medical image segmentation built upon the SAM-2 architecture. Specifically, we introduce a temporal adapter into the image encoder to capture image correlations and a confidence-driven memory mechanism to store high-certainty features for later retrieval. This memory-based strategy counters the pervasive noise in large-scale medical datasets and mitigates catastrophic forgetting when encountering new tasks or modalities. To train and evaluate SAMed-2, we curate MedBank-100k, a comprehensive dataset spanning seven imaging modalities and 21 medical segmentation tasks. Our experiments on both internal benchmarks and 10 external datasets demonstrate superior performance over state-of-the-art baselines in multi-task scenarios. The code is available at: \url{https://github.com/ZhilingYan/Medical-SAM-Bench}.


\end{abstract}
%
%
%
\section{Introduction}

Medical image segmentation plays a pivotal role in clinical practice, supporting disease diagnosis, surgical planning, and treatment evaluation~\cite{liu2021review}. Traditional convolutional neural networks (CNNs)~\cite{o2015introduction}, such as U-Net~\cite{ronneberger2015u} and its variants~\cite{isensee2021nnu,cao2022swin}, have proven effective in specific tasks and single imaging modalities, but they typically require large labeled datasets and substantial retraining for each new application. This constraint often hinders practical deployment, given the high cost of data annotation and the diversity of medical imaging modalities.

To overcome these limitations, MedSAM~\cite{ma2024segment} was proposed as the first segmentation foundation model in the medical domain, demonstrating promising zero-shot performance on various medical tasks. However, as a foundation model, MedSAM does not fully address several inherent challenges in the medical domain. First, medical data—such as contiguous CT/MR slices or surgery videos—requires effective integration of temporal information. Second, the diverse and noisy nature of medical images, marked by various perturbations~\cite{goyal2018noise}, makes it difficult to maintain consistent segmentation quality across heterogeneous datasets. Third, the necessity for continuous learning across different organs and modalities introduces the risk of catastrophic forgetting~\cite{kumari2023continual,wang2024comprehensive}, where previously learned knowledge deteriorates over time.

Meanwhile, SAM-2~\cite{ravi2024sam} in the natural image domain utilizes a memory base to capture sequential information, offering a viable solution for temporal modeling. However, simply fine-tuning SAM-2 on medical datasets addresses only part of the challenge. It does not fully resolve the problems of noise robustness and knowledge retention when facing multi-task or multi-modality scenarios.

In this paper, we introduce SAMed-2, a new foundation model for medical image segmentation that extends SAM-2 with two innovations. First, we incorporate a temporal adapter in the image encoder to exploit temporal correlations. Second, we propose a confidence-driven memory mechanism that selectively stores high-confidence features during training and retrieves them by similarity for inference, effectively mitigating noise and forgetting. Additionally, we curate a large-scale dataset, MedBank-100k, covering diverse modalities and anatomies, to benchmark our method. Our contributions can be summarized as follows: (1) we propose SAMed-2, a dedicated foundation model for medical image segmentation; (2) we design a confidence-driven memory mechanism to handle noisy data and alleviate catastrophic forgetting; (3) we curate MedBank-100k, a large-scale dataset that spans multiple imaging modalities and anatomical regions; and (4) our extensive experiments demonstrate that SAMed-2 achieves state-of-the-art performance on both internal and external tasks, notably improving external zero-shot results by 10.53\%.

\section{Related Work}

In natural image segmentation, foundation models generalize impressively across diverse tasks~\cite{kirillov2023segment,ravi2024sam,zou2023segment} by leveraging massive datasets and scalable architectures to ``segment anything'' with appropriate prompts. However, direct application to medical imaging is challenging due to significant domain gaps. To address this, specialized adaptations have emerged~\cite{dai2023samaug,zhang2023towards}, with SAMed~\cite{zhang2023customized} and MA-SAM~\cite{chen2024ma} extend SAM for volumetric data, while PolypSAM~\cite{li2024polyp} focuses on polyp detection. Although these methods validate SAM-like approaches in medicine, they are typically tailored to specific tasks. MedSAM-2~\cite{zhu2024medical} offers a more general solution by uniformly handling 2D and 3D data via a memory base; however, it ignores temporal information and its memory module only serves the current task, limiting its ability to retain knowledge for evolving modalities or anatomical targets. This underscores the need for stronger multi-task adaptability. In contrast, our work addresses these shortcomings by explicitly incorporating temporal information and employing a confidence-driven memory mechanism, ensuring robust knowledge retention and enhanced multi-task adaptability.

\section{Method}

\begin{figure}[h]
\includegraphics[width=\textwidth]{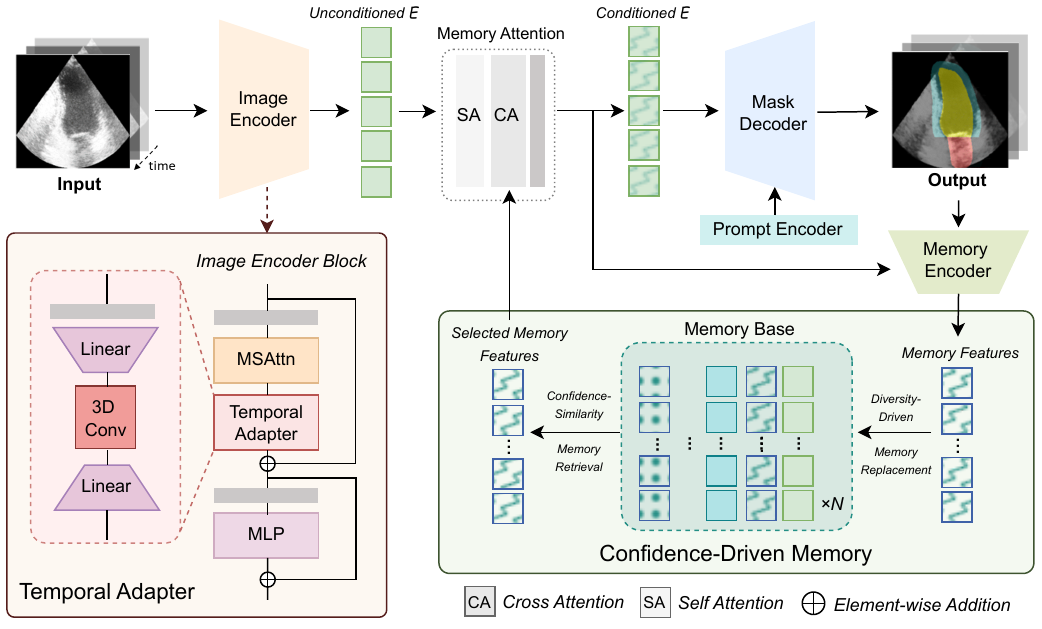}
\caption{Workflow of SAMed-2. It integrates a temporal adapter in the image encoder to capture multi-dimensional context and a confidence-driven memory module to store high-certainty features. During inference, the model retrieves these memory features and fuses them with image embeddings via attention.} \label{fig_method}
\end{figure}

In this section, we first briefly introduce an overview of the SAM-2 architecture, then describe our strategy for integrating volumetric or temporal information, and finally detail the proposed confidence-driven memory mechanism. An overview of the framework is illustrated in Fig.~\ref{fig_method}.

\subsection{Overview of SAM-2}

For SAM-2, given an input image \(I \in \mathbb{R}^{H \times W \times C}\), a ground-truth mask \(M\), and the IoU confidence \(y\), the goal is to predict a segmentation mask \(\hat{M}\) and an IoU score \(\hat{y}\). First, \(I\) is encoded by image encoder \(\mathcal{E}_{\text{img}}\) into an unconditioned feature \(E\). Memory attention \(\mathcal{A}\) then fuses \(E\) with previous entries from the memory base \(\mathcal{M}\) to produce a conditioned embedding \(E_{\text{cond}}\). Meanwhile, the prompt encoder \(\mathcal{E}_{\text{prompt}}\) transforms user prompts into an embedding \(P\). Finally, the mask decoder \(\mathcal{D}\) takes \(E_{\text{cond}}\) and \(P\) to predict \(\hat{M}\) and \(\hat{y}\), while the memory encoder \(\mathcal{E}_{\text{mem}}\) downsamples \(\hat{M}\) to update \(\mathcal{M}\).

\subsection{Temporal Adapter}
Leveraging volumetric/temporal knowledge is crucial for transferring SAM-2 to medical imaging. Inspired by~\cite{chen2024ma}, we integrate spatial attention with a temporal adapter in SAM-2 to fuse spatial and temporal information (Fig.~\ref{fig_method}). Specifically, let $\mathbf{x} \in \mathbb{R}^{B \times H \times W \times C}$ be the block input, each block does: 

\begin{equation}
\mathbf{x}_{\text{out}} 
= \mathbf{x} + \text{DropPath}\Bigl(
    \text{TemporalAdapter}\bigl(
        \text{MultiHeadAttn}(\text{LN}(\mathbf{x}))
    \bigr)
\Bigr)
\end{equation}
\setlength{\belowdisplayskip}{10pt}
which is followed by a MLP layer with residual connection.
For $\text{TemporalAdapter}(\cdot)$, it consists of a normalization layer, a linear down-projection layer, a 3D convolution and a linear up-projection layer, expressed as:
\begin{equation}
    \mathbf{x}_{\text{temp}} = \mathbf{x}_{\text{attn}} + W_{\text{up}} \left( \sigma \left( \text{Conv3D} \left( W_{\text{down}} \, \text{LN} (\mathbf{x}_{\text{attn}}) \right) \right) \right).
\end{equation}
\setlength{\belowdisplayskip}{5pt}
where the 3D convolution captures volumetric or temporal information, the down-projection and up-projection layers help control parameter overhead while maintaining compatibility with the SAM-2 pipeline. 


\subsection{Confidence-Driven Memory Mechanism}

Large-scale medical pre-training is prone to noise and uncertainty, and training on diverse modalities can lead to catastrophic forgetting. To address them, we propose a confidence-driven memory mechanism that selectively stores high-confidence features in pre-training and retrieves them by similarity in inference.

Specifically, we build a memory base
$\mathcal{M} = \{(F_i, PE_i, \hat{y}_i, E_i)\}_{i=1}^{N}$, where \(F_i \in \mathbb{R}^{C \times H \times W}\) is mask feature, \(PE_i\) is positional encoding, \(\hat{y}_i\) is IoU confidence, and \(E_i\) is an unconditioned image embedding for \(i\)th frame. When a new frame arrives with $(F_\text{new}, PE_\text{new}, \hat{y}_\text{new}, E_\text{new})$, the mechanism proceeds as follows:

\subsubsection{Confidence-Similarity Memory Retrieval.}
To select the $K$ most relevant memory entries with high confidence 
$\{\,\pi(1),\;\pi(2),\;\dots,\;\pi(K)\}$:
\begin{equation}
\pi \;=\; \operatorname{argsort}\Bigl(\bigl\{\,s_i \;+\; \sigma(\hat{y}_i)\bigr\}_{i=1}^N\Bigr).
\end{equation}
where $s_i$ is the cosine similarity:
\begin{equation}
s_i = \cos(\theta) = \frac{E_i^{T} E_{\text{new}}}{\|E_i\|_2 \|E_{\text{new}}\|_2}.
\end{equation}

The memory features $\{(F_i, PE_i)\}_{i \in K}$ are then fed into the memory attention $\mathcal{A}$ to get the conditioned feature embedding $E_{\text{cond}}$ via:
\begin{equation}
    E_{\text{cond}} = \mathcal{A} \left( E_{\text{new}}, PE_{\text{new}}, \{ F_i, PE_i \}_{i \in K} \right).
\end{equation}
\subsubsection{Confidence-Driven Memory Replacement.}

During pre-training, some memory entries may become unreliable or redundant. Hence, whenever a new frame arrives, we attempt to replace an existing memory item if the new item is sufficiently confident.
Suppose our memory base $\mathcal{M} = \{(F_i, PE_i, \hat{y}_i, E_i)\}_{i=1}^{N}$.
Firstly, we compute similarity of the new item to each stored entry $F_i$ and find the memory entry that has the highest similarity:
\begin{equation}
    s_ \text{max}= \arg\max_i \left( \frac{F_i^{T} F_{\text{new}}}{\|F_i\|_2 \|F_{\text{new}}\|_2} \right).
\end{equation}

Then replace the corresponding memory entry if $\hat{y}_{i,s_\text{max}} < \hat{y}_{\text{new}}$, which means the new item is sufficiently confident:
\begin{equation}
\begin{aligned}
\mathcal{M}
&=
\Bigl(\mathcal{M}\setminus\{F_{i,s_{\text{max}}}, PE_{i,s_{\text{max}}}, \hat{y}_{i,s_{\text{max}}}, E_{i,s_{\text{max}}}\}\Bigr)\\
&\quad \cup \Bigl\{F_{\mathrm{new}}, PE_{\mathrm{new}}, \hat{y}_{\mathrm{new}}, E_{\mathrm{new}}\Bigr\}.
\end{aligned}
\end{equation}

This ensures that uncertain entries are excluded, while explicitly retaining previous features over time, thereby mitigating forgetting issues.

\section{Experiment}

\subsection{MedBank-100k: Equipping SAM-2 with Medical Knowledge}

To adapt SAM-2 for biomedical tasks, we built MedBank-100k
, a comprehensive dataset from public sources that covers 21 segmentation tasks and seven imaging modalities (Table~\ref{tab_data}), totaling 122,594 frame-mask pairs. Given the diversity of modalities, formats, and shapes, we standardized and normalized the data via the following pre-processing steps~\cite{ma2024segment,huang2024segment}: (1) For video data, only frames with nonzero label sums (i.e., meaningful annotations) are retained; (2) 2D images are randomly shuffled while preserving temporal/spatial relationships of video/3D slices; (3) Images whose shortest edge is less than half the longest edge are discarded to prevent excessive blurring; (4) Multi-class masks are separated by class. These steps ensure uniformity and compatibility across all images and videos, which is essential for effective training and evaluation.

\begin{minipage}{0.48\textwidth}
  \centering
  \scalebox{0.8}{
  \begin{tabular}{l r r}
    \toprule
    Modality          & \# Task & \# Image \\
    \midrule
    Fundus      & 1       & 559 \\
    Dermoscopy        & 1       & 2621 \\
    X-Ray             & 1       & 23822 \\
    CT                & 10      & 34521 \\
    MR                & 6       & 19522 \\
    Colonoscopy       & 1       & 3838 \\
    Echocardiography  & 1       & 1800 \\
    Others            & -       & 35911 \\
    \bottomrule
  \end{tabular}}
  \captionof{table}{Overview of MedBank-100k’s distribution by imaging modality.}
  \label{tab_data}
\end{minipage}
\hfill
\begin{minipage}{0.46\textwidth}
  \centering
  \scalebox{0.8}{
  \includegraphics[width=0.9\textwidth]{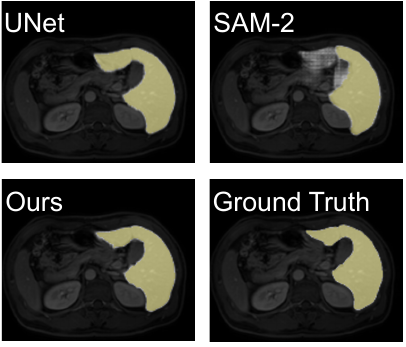}}
  \captionof{figure}{Qualitative visualization of segmentation results.}
  \label{fig_vis}
\end{minipage}

\begin{table*}[h]
\vspace{-5pt}
  \centering
  \setlength{\tabcolsep}{3pt}      
  \renewcommand{\arraystretch}{1.2} 
  
  \resizebox{0.87\textwidth}{!}{%
    \begin{tabular}{l |l |ccccc |c}
      \hline
      Task         & Modality       & \textbf{Ours}  & MedSAM-2 & SAM2  & MedSAM & SAM   & U-Net  \\
      \hline
      Thyroid~\cite{eiraoi_thyroidultrasound}      & US     & 0.8663 & 0.6756 & 0.4967 & \cellcolor{lightgray}0.8863 & 0.8655 & 0.8052 \\
      OpticCup~\cite{sivaswamy2014drishti}     & Fundus   & \cellcolor{lightgray}0.8936 & 0.3820 & 0.5589 & 0.8823 & 0.6064 & 0.8833 \\
      Melanoma~\cite{gutman2016skin}     & Derm     &  0.9167 & 0.6601 & 0.6158 & \cellcolor{lightgray}0.9364 & 0.8322 & 0.8884 \\
      Hippocampus~\cite{antonelli2022medical}  & MR             &  0.8010 & 0.5650 & 0.5237 & 0.7410 & 0.7196 & \cellcolor{lightgray}0.8097 \\
      Prostate~\cite{antonelli2022medical}     & MR             & \cellcolor{lightgray}0.8601 & 0.7188 & 0.6534 &  0.8546 & 0.8302 & 0.8406 \\
      BrainTumor~\cite{antonelli2022medical}   & MR             & \cellcolor{lightgray}0.6726 & 0.5852 & 0.4317 & 0.5925 & 0.5618 &  0.6589 \\
      Liver~\cite{antonelli2022medical}        & CT             & \cellcolor{lightgray}0.7738 & 0.6996 & 0.4457 &  0.7636 & 0.6771 & 0.7501 \\
      LiverTumor~\cite{antonelli2022medical}   & CT             &  0.4295 & 0.3996 & 0.2047 & 0.4192 & 0.3914 & \cellcolor{lightgray}0.4914 \\
      Spleen~\cite{antonelli2022medical}       & CT             &  0.8813 & 0.7732 & 0.7006 & 0.8789 & 0.8650 & \cellcolor{lightgray}0.8937 \\
      ColonTumor~\cite{antonelli2022medical}   & CT             & \cellcolor{lightgray}0.7005 & 0.6306 & 0.2464 &  0.6955 & 0.6432 & 0.6901 \\
      \hline
      \multicolumn{2}{l|}{\textbf{Average}}  & \cellcolor{lightgray}0.6938 & 0.5796 & 0.4375 & 0.6277 & 0.5958 &  0.6879 \\
      \hline
    \end{tabular}%
  }
  \caption{External validation of our method and state-of-the-art methods. *US: Ultrasound, Derm: Dermoscopy. \textit{Note}: We consistently highlight the best performances in each row with \colorbox{lightgray}{Best} for all subsequent tables.}
  \label{tab_external}
\end{table*}

\subsection{Implementation Setting}

The dataset is randomly split into training and test sets (9:1) at the image level to prevent potential data leakage. Our model is initialized with the pre-trained SAM-2 (Hiera-S variant) and optimized using AdamW~\cite{loshchilov2017decoupled} with an initial learning rate of 1e-4; the loss follows SAM-2~\cite{ravi2024sam}. Experiments run for 100 epochs on a single NVIDIA H100 GPU, with the best checkpoint selected as final. For fair comparison, all methods share the same settings. We compare our model against SAM~\cite{kirillov2023segment}, SAM-2~\cite{ravi2024sam}, MedSAM~\cite{ma2024segment}, MedSAM-2~\cite{zhu2024medical}, and U-Net~\cite{ronneberger2015u}, where U-Net is trained per task and others use official checkpoints with bounding boxes as prompts following~\cite{ma2024segment}. The evaluation metric is the Dice Similarity Coefficient (DSC), where higher values are better.

\begin{table*}[h]
  \centering
  \setlength{\tabcolsep}{3pt}      
  \renewcommand{\arraystretch}{1.2} 
  
  \resizebox{0.9\textwidth}{!}{ 
    \begin{tabular}{l|l|ccccc|c}
      \hline
      Task & Modality & \textbf{Ours} & MedSAM-2 & SAM2 & MedSAM & SAM & U-Net \\ \hline
      OpticCup            & Fundus     & \cellcolor{lightgray}0.8971   & 0.4040   & 0.6209   &  0.8643   & 0.6230   & 0.8073 \\ 
      Melanoma            & Derm       & \cellcolor{lightgray}0.9119   & 0.6877   & 0.5293   &  0.9106   & 0.8398   & 0.8169 \\ 
      Lung                & X-Ray            &  0.9007   & 0.6110   & 0.6668   & 0.9001   & 0.6840   & \cellcolor{lightgray}0.9700 \\ 
      Spleen              & CT               &  0.8566   & 0.8260   & 0.4441   & \cellcolor{lightgray}0.8755   & 0.8402   & 0.5239 \\ 
      Esophagus           & CT               & \cellcolor{lightgray}0.6881   & 0.5670   & 0.1139   & 0.6672   &  0.6879   & 0.3464 \\ 
      Liver               & CT               &  0.8332   & 0.8159   & 0.6752   & 0.8330   & 0.7935   & \cellcolor{lightgray}0.8730 \\ 
      Stomach             & CT               & \cellcolor{lightgray}0.8347   & 0.7131   & 0.4604   &  0.8325   & 0.8039   & 0.4383 \\ 
      InferiorVenaCava    & CT               & \cellcolor{lightgray}0.7800   & 0.6821   & 0.1738   &  0.7799   & 0.7497   & 0.5471 \\ 
      Pancreas            & CT               & \cellcolor{lightgray}0.5785   & 0.5537   & 0.1343   &  0.5775   & 0.5542   & 0.2569 \\ 
      AdrenalGland        & CT               & \cellcolor{lightgray}0.2886   & 0.2838   & 0.0194   &  0.2853   & 0.1976   & 0.1869 \\ 
      Liver               & MR               & \cellcolor{lightgray}0.8779   & 0.7627   & 0.6895   &  0.8773   & 0.8313   & 0.8748 \\ 
      Aorta               & MR               &  0.9033   & 0.7016   & 0.5158   & 0.9029   & 0.9012   & \cellcolor{lightgray}0.9315 \\ 
      InferiorVenaCava    & MR               & \cellcolor{lightgray}0.8007   & 0.6454   & 0.3014   & \cellcolor{lightgray}0.8007   &  0.8006   & 0.6742 \\ 
      Gallbladder         & MR               & 0.8256   & 0.6684   & 0.3602   &  0.8436   & \cellcolor{lightgray}0.8450   & 0.6879 \\ 
      Esophagus           & MR               &  0.6402   & 0.5075   & 0.2865   & 0.6401   & 0.6395   & \cellcolor{lightgray}0.6453 \\ 
      Stomach             & MR               & \cellcolor{lightgray}0.7845   & 0.7287   & 0.5386   &  0.7843   & 0.7540   & 0.7139 \\ 
      KidneyTumor         & CT               & \cellcolor{lightgray}0.8212   & 0.7211   & 0.4182   &  0.8210   & 0.8122   & 0.6486 \\ 
      LiverTumor          & CT               &  0.4443   & 0.4101   & 0.2102   & 0.4440   & 0.3837   & \cellcolor{lightgray}0.4528 \\ 
      LungTumor           & CT               & \cellcolor{lightgray}0.6709   & 0.5225   & 0.3204   &  0.6707   & 0.6618   & 0.5492 \\  \hline
        \multicolumn{2}{l|}{\textbf{Average}} & \cellcolor{lightgray}0.7086   & 0.6243   & 0.3972   &  0.7076   & 0.6730   & 0.6516 \\ \hline
      Polyp*               & Colon     & \cellcolor{lightgray}0.8183   & 0.6966   & 0.3535   &  0.8181   & 0.7416   & 0.2391 \\ 
      LV$_{epi}$*               & Echo             &  0.5733   & 0.4983   & 0.4404   & 0.3412   & 0.1667   & \cellcolor{lightgray}0.7914 \\ \hline
      \multicolumn{2}{l|}{\textbf{Average}} & \cellcolor{lightgray}0.7118   & 0.6247   & 0.3954   &  0.6998   & 0.6549   & 0.6221 \\ \hline
    \end{tabular}
  }
  \caption{Internal validation of our method and state-of-the-art methods. *Derm: Dermoscopy, Colon: Colonoscopy, Echo: Echocardiography.}
  \label{tab_internal}
\end{table*}

\subsection{Quantitative analysis}

We evaluated our model with external and internal validation. For external validation, 10 segmentation tasks (Table~\ref{tab_external}) were used, representing new patients, imaging conditions, and tasks. Our model achieves the best overall DSC (0.6938), ranking first on five tasks and second on the remainder, while U-Net—trained per task—is second overall (0.6879). Notably, our model outperforms MedSAM by 10.53\% on 8/10 tasks. Incorporating medical knowledge improves performance on unseen tasks (MedSAM-2 vs. SAM-2 and MedSAM vs. SAM).

Internal validation involved 21 segmentation tasks (Table~\ref{tab_internal})
. Our method achieves the highest average DSC (0.7118), 1.71\% higher than MedSAM, and the best DSC in 13 of 21 tasks, with competitive performance on the remainder. The improvements of MedSAM-2 over SAM-2 and MedSAM over SAM underscore the gap between natural and biomedical data and the need for adaptation. Fig.~\ref{fig_vis} shows a liver segmentation sample: while compared models under- or over-segments target boundaries, our method accurately segments the liver.

\subsection{Ablation Study}

We performed ablation experiments on internal and external tasks to assess: (1) the effectiveness of each key component; (2) few-shot scaling on external tasks; (3) choice of memory base size; and (4) memory retrieval strategies at inference.

\begin{table*}[h]
    \centering
    \small
    \setlength{\tabcolsep}{6pt}  
    \renewcommand{\arraystretch}{1.2} 

    \caption{Ablation on each key component in our method.}

    \resizebox{0.85\textwidth}{!}{  
    \begin{tabular}{l cccc}  
        \toprule
        \multirow{2}{*}{\textbf{Method}} & \multicolumn{2}{c}{Internal Validation} & \multicolumn{2}{c}{External Validation} \\
        \cmidrule(lr){2-3} \cmidrule(lr){4-5}
        & Spleen & InferiorVenaCava & Liver & ColonTumor \\
        \midrule
        Baseline & 0.7656 & 0.7062 & 0.7000 & 0.6422 \\
        \texttt{+} Memory Retrieval & 0.7716 & 0.7047 & 0.7001 & 0.6491 \\
        \texttt{+} Confidence & 0.7850 & 0.7106 & 0.7025 & 0.6491 \\
        \texttt{+} Temporal Adapter &  0.8448 &  0.7668 &  0.7680 &  0.6878 \\
        \textbf{Ours} & \cellcolor{lightgray}0.8566 & \cellcolor{lightgray}0.7800 & \cellcolor{lightgray}0.7738 & \cellcolor{lightgray}0.7005 \\
        \bottomrule
    \end{tabular}
    \label{tab_ablation_compo}
    }
\end{table*}

\begin{figure}[h]
    \centering
    \begin{minipage}{0.48\textwidth}
        \centering
        \includegraphics[width=\textwidth]{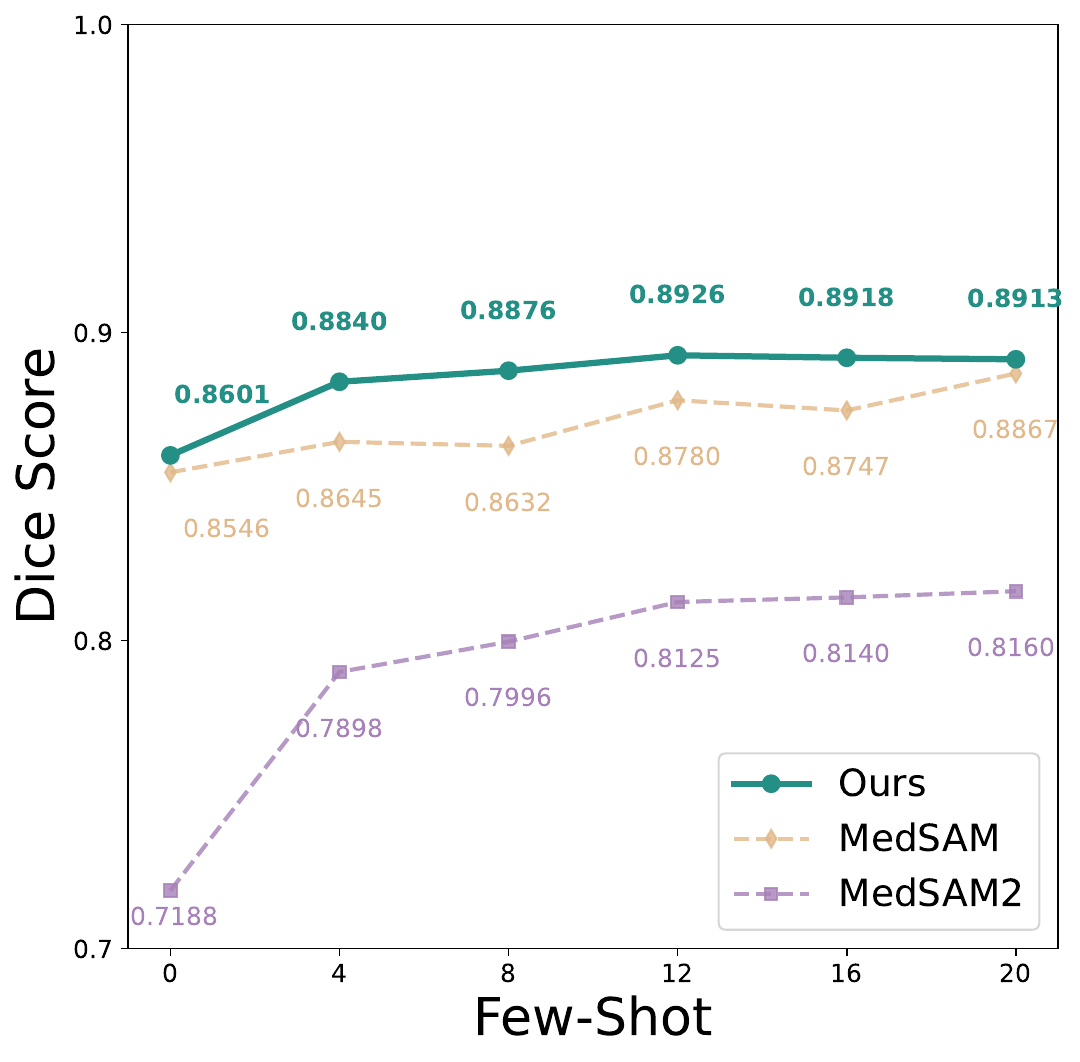}
        \caption{Few-shot scaling result of our method and compared methods on external prostate segmentation task.}
        \label{fig_few_shot}
    \end{minipage}
    \hfill
    \begin{minipage}{0.48\textwidth}
        \centering
        \includegraphics[width=0.97\textwidth]{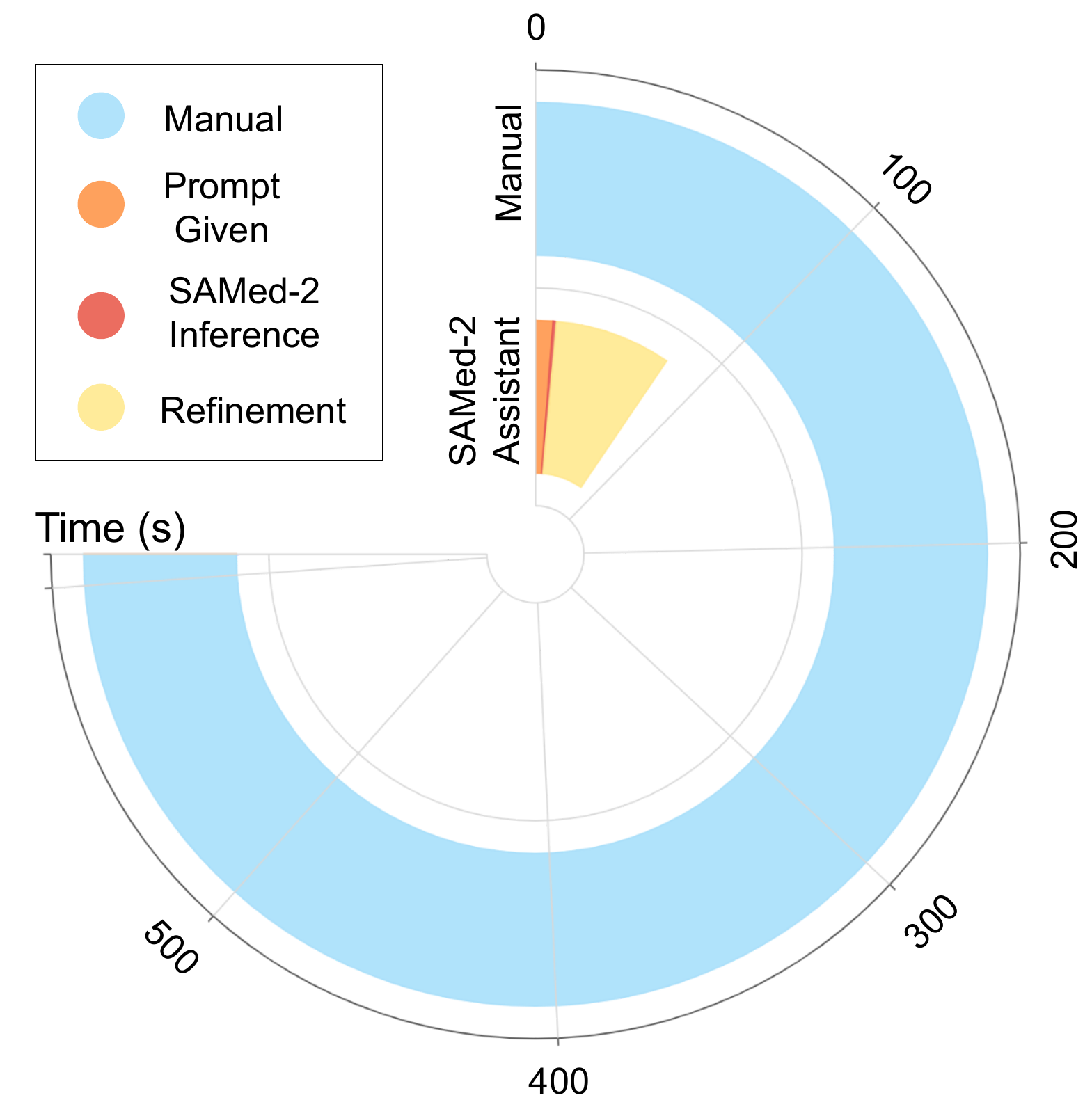}
        \caption{Annotation time comparison between manual annotation and SAMed-2-assisted annotation.}
        \label{fig_human_user}
    \end{minipage}
\end{figure}

\paragraph{1) Effectiveness of each component:}
We evaluated key components against a baseline without these modules, \textit{i.e.}, memory retrieval, confidence in memory, and temporal adapter, denoted as ``Memory Retrieval'', ``Confidence'', and ``Temporal Adapter'' in Table~\ref{tab_ablation_compo}. The ``Baseline'' is SAM-2 pretrained on medical data. Comparing the baseline with the model using temporal adapter yields improvements up to 10.34\% on internal spleen segmentation and 9.71\% on external liver segmentation, underscoring the importance of temporal information for multi-dimensional volumes. Moreover, integrating memory retrieval and confidence boosts performance: the confidence-driven memory mechanism (last row) outperforms the model with only a temporal adapter (fourth row) on internal and external tasks, demonstrating the necessity of explicitly storing prior knowledge and filtering uncertain samples to reduce noise.
\paragraph{2) Few-shot scaling:}
Different from zero-shot validation, we tested external tasks with limited training data. As Fig.~\ref{fig_few_shot} shows, SAMed-2 outperforms other methods for 0–20 shots. It maintains strong performance with as few as 4 shots, demonstrating the generality of our pre-training—and further improves with more shots. This underscores its excellent transferability and scalability in low-label settings, which is crucial in clinical scenarios where annotations are scarce.

\begin{table*}[h]
    \centering
    \small
    \setlength{\tabcolsep}{6pt}      
    \renewcommand{\arraystretch}{1.2} 

    \caption{Ablation study on confidence-driven memory mechanism.}
    \resizebox{0.87\textwidth}{!}{  
    \begin{tabular}{l cccc}
        \hline
        \multirow{2}{*}{Variant} & \multicolumn{2}{c}{Internal Validation} & \multicolumn{2}{c}{External Validation} \\
        \cmidrule(lr){2-3} \cmidrule(lr){4-5}
        & Spleen & InferiorVenaCava & Liver & ColonTumor \\
        \hline
        \hline
        \multicolumn{5}{c}{Memory Base Size} \\        
        \hline
        \hline
        0   &  0.8558  & 0.7651 & 0.7677 &  0.6865 \\
        16  & 0.8440  &  0.7656 &  0.7697 & 0.6832 \\
        640 & \cellcolor{lightgray}0.8566  & \cellcolor{lightgray}0.7800 & \cellcolor{lightgray}0.7738 & \cellcolor{lightgray}0.7005 \\
        \hline
        \hline
        \multicolumn{5}{c}{Memory Retrieval Approach} \\
        \hline
        \hline
        Random &  0.8527  & 0.7602 & 0.7674 & 0.6815 \\
        Confidence-Similarity & \cellcolor{lightgray}0.8566  & \cellcolor{lightgray}0.7800 & \cellcolor{lightgray}0.7738 & \cellcolor{lightgray}0.7005 \\
        \hline
    \end{tabular}
    }
    \label{tab_memory}
\end{table*}

\paragraph{3) Choice of memory base sizes:}
We evaluated the effect of memory base size during inference. A size of ``0'' indicates no pre-trained memory is used and the validation task’s own memory guides prediction, while ``16'' and ``640'' denote pre-trained memory bases, with “640” as our default. As shown in the upper part of Table~\ref{tab_memory}, “0” and “16” yield comparable performance. A small memory base may store incomplete or mismatched features that conflict with the model’s learned representations, whereas a large memory base enhances performance. This underscores the importance of memory diversity for multi-task learning and validates our mechanism.

\paragraph{4) Choice of memory retrieval approaches:}
When retrieving features from the memory base during inference, we used both confidence and similarity between memory features and current image embeddings. As shown in the lower part of Table~\ref{tab_memory}, our ``Confidence-Similarity'' retrieval outperforms random retrieval on both internal and external tasks, confirming its noise robustness.

\subsection{Human User Study}

We conducted a human user study on an in-house CMR dataset (unseen by SAMed-2) to assess annotation efficiency. A cardiovascular imaging expert (>8 years' experience) annotated data manually and with SAMed-2 assistance. As shown in Fig.~\ref{fig_human_user}, manual annotation averages $609\,\mathrm{s} \pm 137.4\,\mathrm{s}$ per frame, while SAMed-2 reduces this to $75.43\,\mathrm{s} \pm 18.16\,\mathrm{s}$—an 87.61\% reduction. These results highlight SAMed-2’s potential to significantly streamline medical segmentation.

\section{Conclusion}

In this paper, we presented SAMed-2, a foundation model for medical image segmentation that extends SAM-2 with temporal awareness and a confidence-driven memory mechanism. By embedding a temporal adapter within the encoder, the model harnesses inter-slice or inter-frame correlations, which proves essential for multi-dimensional imaging tasks. The confidence-based memory module, trained with a large-scale yet heterogeneous dataset (MedBank-100k), selectively stores and retrieves high-certainty features, enhancing robustness to noise and mitigating catastrophic forgetting in multi-task and multi-modality settings. These results confirm that SAMed-2 effectively bridges the gap between ``segment anything'' objectives and the unique demands of clinical imaging.

\bibliographystyle{splncs04}
\bibliography{mybibliography}




\end{document}